\documentclass{article}
\usepackage{nips14submit_e,times}
\usepackage{latexsym}
\usepackage{amsmath}
\usepackage{url}
\usepackage{graphicx}
\usepackage{haldefs}

\title{Dynamic Feature Scaling for \\ Online Learning of Binary Classifiers}

\author{
Danushka~Bollegala \\
University of Liverpool, \\
Liverpool, \\United Kingdom. \\
\texttt{danushka.bollegala@liverpool.ac.uk}} 

\nipsfinalcopy 

\begin{document}

\maketitle

\begin{abstract}
Scaling feature values is an important step in numerous machine learning tasks.
Different features can have different value ranges and some form of a feature scaling is often
required in order to learn an accurate classifier. 
However, feature scaling is conducted as a preprocessing task prior to learning. 
This is problematic in an online setting because of two reasons. 
First, it might not be possible to accurately determine the value range of a feature at the
initial stages of learning when we have observed only a few number of training instances.
Second, the distribution of data can change over the time, which render obsolete any feature scaling
that we perform in a pre-processing step.
We propose a simple but an effective method to dynamically scale features at train time, 
thereby quickly adapting to any changes in the data stream.
We compare the proposed dynamic feature scaling method against more complex methods
for estimating scaling parameters using several benchmark datasets for binary classification.
Our proposed feature scaling method consistently outperforms more complex methods on all
of the benchmark datasets and improves classification accuracy of a state-of-the-art online binary classifier algorithm.
\end{abstract}

\section{Introduction}
\label{sec:intro}

Machine learning algorithms require train and test instances to be represented
using a set of features. For example, in supervised document classification \cite{Crammer:2006},
 a document is often represented as a vector of its words and the value of a feature is set to the
number of times the word corresponding to the feature occurs in that document.
However, different features occupy different value ranges, and often one must scale
the feature values before any supervised classifier is trained.
In our example of document classification, there are both highly frequent words
(e.g. stop words) as well as extremely rare words. Often, the relative difference of a value of
a feature is more informative than its absolute value. Therefore, feature scaling
has shown to improve performance in classification algorithms.

Typically, feature values are scaled to a standard range in a preprocessing step before 
using the scaled features in the subsequent learning task. However, this preprocessing approach
to feature value scaling is problematic because of several reasons. First, often feature scaling
is done in an unsupervised manner without consulting the labels assigned to the training
instances. Although this is the only option in unsupervised learning tasks such as document
clustering, for supervised learning tasks such as document classification,
where we do have access to the label information, we can use 
the label information also for feature scaling.
Second, it is not possible to perform feature scaling as a preprocessing step in \textit{one-pass}
online learning setting. In one-pass online learning we are allowed to traverse through the
set of training instances only once. Learning from extremely large datasets such as twitter
streams or Web scale learning calls for algorithms that require only a single pass over the
set of training instances. In such scenarios it is not possible to scale the 
feature values beforehand by using statistics from the entire training set.
Third, even if we pre-compute scaling parameters for a feature, those values might become
obsolete in an online learning setting in which the statistical properties of the training instances
vary over the time. For example, a twitter text stream regarding a particular keyword
might change overtime and the scaling factors computed using old data might not be
appropriate for the new data.

We study the problem of dynamically scaling feature values at run time for online learning.
The term \textit{dynamic feature scaling} is used in this paper to refer to the practice of
scaling feature values at run time as opposed to performing feature scaling as a pre-processing step
that happens prior to learning. We focus on binary classifiers as a specific example. 
However, we note that the proposed method can be easily extended to multi-class classifiers. 
We propose two main approaches for dynamic feature scaling in this paper:
(a) \textit{Unsupervised Dynamic Feature Scaling} (Section \ref{sec:unsupscale}), in which we do not consider the label information
assigned to the training instances for feature scaling, and 
(b) \textit{Supervised Dynamic Feature Scaling} (Section \ref{sec:supscale}), in which we consider the label information
assigned to the training instances for feature scaling.

All algorithms we propose in this paper can be trained under the one-pass online learning setting, where only
a single training instance is provided at a time and only the scale parameters and feature
weights are stored in the memory. This enables the proposed method to
(a) efficiently adapt to the varying statistics in the data stream,
(b) compute the optimal feature scales such that the likelihood of the training data 
under the trained model is maximized,
and (c) train from large datasets where batch learning is impossible because of memory requirements.
We evaluate the proposed methods in combination with different online learning algorithms
using three benchmark datasets for binary classification.
Our experimental results show that, interestingly, the much simpler unsupervised dynamic feature scaling
method consistently improves all of the online binary classification algorithms we compare, including the state-of-the-art
classifier of \cite{Crammer:2006}.

\section{Related Work}
\label{sec:related}

Online learning has received much attention lately because of the necessity to learn from large training datasets
such as query logs in a web search engine \cite{pantel-lin-gamon:2012:ACL2012}, 
web-scale document classification or clustering \cite{Madani:2010}, and sentiment analysis
on social media \cite{Jiang:ACL:2011,Dredze:EMNLP:2010}. 
Online learning toolkits that can efficiently learn from large datasets are made available 
such as Vowpal Wabbit\footnote{\url{https://github.com/JohnLangford/vowpal_wabbit}}
and OLL\footnote{\url{https://code.google.com/p/oll/}} (Online Learning Library).
Online learning approaches are attractive than their batch learning counterparts when the training data involved
is massive due to two main reasons. First, the entire dataset might not fit into the main memory of a single computer
to perform a batch optimization. Although there has been some recent progress in distributed learning algorithms 
\cite{Gopal:ICML:2013,Duchi:NIPS:2010,Liu:WWW:2010}
that can distribute the batch optimization process across a series of machines, setting up and debugging such
a distributed learning environment remains a complex process. On the other hand, online learning algorithms consider only
a small batch (often referred to as a \textit{mini batch} in the literature) or in the extreme case a single training
instance. Therefore, the need for large memory spaces can be avoided with online learning.
Second, a batch learning algorithm requires at least one iteration over the entire dataset to produce a classifier.
This can be time consuming for large training datasets. On the other hand, online learning algorithms
can produce a relatively accurate classifier even after observing a handful of training instances.

Online learning is a vast and active research field and numerous algorithms have been proposed
in prior work to learn classifiers
 \cite{Crammer:EMNLP:2019,Crammer:NIPS:2008,Mejer:EMNLP:2010,Mejer:2011,Duchi:COLT:2010,Ma:AISTAT:2010}.
A detailed discussion of online classification algorithms is beyond the scope of this paper.
Some notable algorithms are the passive-aggressive (PA)
algorithms \cite{Crammer:2006}, confidence-weighted linear classifiers \cite{Dredze:ICML:2008} and
their multi-class variants \cite{Crammer:EMNLP:2019,Crammer:NIPS:2008}.
In passive-aggressive learning, the weight vector for the binary classifier is updated only when
a misclassification occurs. If the current training instance can be correctly classified using the current
weight vector, then the weight vector is not updated. In this regard, the algorithm is considered \textit{passive}.
On the other hand, if a misclassification occurs, then the weight vector is \textit{aggresively} updated such that it can correctly
classify the current training instance with a fixed margin. Passive-aggressive algorithm has consistently outperformed
numerous other online learning algorithms across a wide-range of tasks. Therefore, it is considered as a state-of-the-art
online binary classification algorithm.  As we demonstrate later, the unsupervised dynamic feature scaling method proposed in this paper
further improves the accuracy of the passive-aggressive algorithm.
Moreover, active-learning \cite{Dredze:ACL:2008} and transfer learning \cite{Zhao:ICML:2011}
approaches have also been proposed for online classifier learning.

One-Pass Online Learning (\textbf{OPOL}) (also known as \textit{stream learning}) \cite{Dredze:EMNLP:2010}
is a special case of online learning in which \textit{only a single-pass is allowed over the set of train instances} by the learning algorithm.
Typically, an online learning algorithm requires multiple passes over a training dataset to reach a convergent point.
 This setting can be considered as an extreme case where
the train batch size is limited to only one instance. 
The OPOL setting is more restrictive than the classical online learning setting where a learning
algorithm is allowed to traverse multiple times over the training dataset. However, OPOL becomes the
only possible alternative in the following scenarios.

\begin{enumerate}
\item The number of instances in the training dataset is so large that it is impossible to traverse
multiple times over the dataset.

\item The dataset is in fact a stream where we encounter new instances continuously. 
For example, consider the situation where we want to train a sentiment classifier from tweets.

\item The data stream changes over time. In this case, even if we can store old data instances
they might not be much of a help to predict the latest trends in the data stream.
\end{enumerate}

It must be noted that OPOL is not the only solution for the first scenario where we have a 
large training dataset. One alternative approach is to select a subset of examples from
the dataset at each iteration and only use that subset for training in that iteration.
One promising criterion for selecting examples for training is curriculum learning \cite{Bengio:ICML:2009}.
In curriculum learning, a learner is presented with easy examples first and gradually  with
the more difficult examples. However, determining the criteria for selecting easy examples
is a difficult problem itself, and the criterion for selecting easy examples
might be different from one task to another.
Moreover, it is not clear whether we can select easy examples from the training dataset
in a sequential manner as required by online learning without consulting the unseen training
examples.

The requirement for OPOL ever increases with the large training datasets and data streams
we encounter on the Web such as social feeds. Most online learning algorithms require several
passes over the training dataset to achieve convergence. For example, Passive-Aggressive algorithms \cite{Crammer:2006}
require at least $5$ iterations over the training dataset to converge, 
whereas, for Confidence-Weighted algorithms \cite{Dredze:ICML:2008} the number of
iterations has shown to be less (ca. $2$). Our focus in this paper is not to develop online
learning algorithms that can classify instances with high accuracy by traversing only once
over the dataset, but to study the effect of feature scaling in the OPOL setting.
To this end, we study both an unsupervised dynamic feature scaling method (Section \ref{sec:unsupscale})
and several variants of a supervised dynamic feature scaling methods (Section \ref{sec:supscale}).

\section{Unsupervised Dynamic Feature Scaling}
\label{sec:unsupscale}

In unsupervised dynamic feature scaling, given a feature $x_j$, 
we compute the mean, $\mu (x_j)$ and the standard deviation $\delta (x_j)$ of the
feature and perform an affine transformation as follows,
\begin{equation}
\label{eq:unsup}
x_j' = \frac{x_j - \mu_j}{\delta_j} .
\end{equation}
This scaling operation corresponds to a linear shift of the feature values by the
mean value of the feature, followed up by a scaling by its standard deviation.
From a geometric point of view, this transformation will shift the origin to the
mean value and then scale axis corresponding to the $j$-th feature to unit standard
deviation. It is used popularly in batch learning setting, in which one can compute the
mean and the standard deviation using \textit{all} the training instances in the training
dataset. However, this is not possible in OPOL, in which we encounter only one instance
at a time. However, even in the OPOL setting, we can compute the mean and the standard
deviation on the fly and constantly update our estimates of those values as new training
instances (feature vectors) are observed. The update equations for the mean $m^k_j$ and
the standard deviation $\sqrt{s^k_j / (k - 1)}$ for the $j$-th feature are as follows \cite{Ling:1974,Randall:1983},
\begin{eqnarray}
\label{eq:mu_sigma}
m^k_j &=& m^{k-1}_{j} + \frac{x^k_j - m^{k-1}_{j}}{k} , \\
s^k &=& s^{k-1} + (x^{k}_{j} - m^{k-1}_{j})(x^k_j - m^k_j) .
\end{eqnarray}
We use these estimates for the mean and the standard deviation to scale features in Equation \ref{eq:unsup}.
The mean and standard deviation are updated throughout the training process.

\section{Supervised Dynamic Feature Scaling}
\label{sec:supscale}


We define the task of supervised dynamic feature scaling task for binary classification in the OPOL setting as follows.
Given a stream of labeled training instances $(\vec{x}_n,t_n)$, in which the class label $t_n$ 
of the $n$-th training instance $x_n$, denoted by a feature vector $\vec{x}_n$, is assumed to be
either $+1$ (positive class) or $-1$ (negative class).
Furthermore, let us assume that the feature space is $M$ dimensional and the value of the $i$-th feature of the
$n$-th instance in the training data stream is denoted by $x_i^n$. In this paper, we 
consider only real-valued features (i.e. $x_i^n \in \R$) because feature scaling is particularly
important for real-valued features.

We define the feature scaling function $\sigma_i(x_i^n)$ for the $i$-th feature as a function
that maps $\R$ to the range $[0,1]$ as follows:
\begin{equation}
\label{eq:trans}
\sigma_i(x_i^n) = \frac{1}{1+\exp(-\alpha_i x_i^n + \beta_i)}. 
\end{equation}
Here, $\alpha_i$ and $\beta_i$ are the scaling parameters for the $i$-th dimension of the 
feature space. Several important properties of the feature scaling function defined by
Equation \ref{eq:trans} are noted. First, the feature transformation function maps all
feature values to the range $[0,1]$ irrespective of the original range in which each feature
value $x_i$ was. For example, one feature might originally be limited to the range $[0,0.001]$,
whereas another feature might have values in the full range of $[0,10000]$. 
By scaling each feature into a common range we can concentrate on the relative values of
those features without being biased by their absolute values. 
Second, the scaling parameters $\alpha_i$ and $\beta_i$ are defined per-feature basis.
This enables us to scale different features using scale parameters appropriate for their
value ranges. Third, the linear transformation $\alpha_i x_i^n - \beta_i$ within
the exponential term of the feature scaling function resembles the typical affine transformations
performed in unsupervised feature scaling. For example, assuming the mean and the standard
deviation of the $i$-th feature to be respectively $\mu_i$ and $\delta_i$, in supervised
classification, features are frequently scaled to $(x_i - \mu_i) / \delta_i$ prior to training 
and testing. The linear transformation within the exponential term in Equation \ref{eq:trans}
can be seen as a special case of this approach with values $\alpha_i = 1 / \delta_i$ and
$\beta_i = \mu_i / \delta_i$.

Then, the posterior probability, $P(t=1 | \vec{x}^n, b, \vec{\alpha}, \vec{\beta})$
of $\vec{x}^n$ belonging to the positive class
is given as follows according to the logistic regression model \cite{PRML}:
\begin{eqnarray}
\label{eq:score}
P(t_n = 1 | \vec{x}^n, b, \vec{\alpha}, \vec{\beta}) = \frac{1}{1+\exp\(-\sum_{i=1}^{M} w_i\sigma_i(x_i^n) - b\)}, \\ \nonumber
P(t_n =1 | \vec{x}^n, b, \vec{\alpha}, \vec{\beta}) = \frac{1}{1+\exp\(-\frac{w_i}{1+\exp(-\alpha_i x_i^n + \beta_i)} -b\)} .
\end{eqnarray}
Here, $w_i$ is the weight associated with the $i$-th feature and $b \in \R$ is 
the bias term. We arrange the weights $w_i$, scaling parameters $\alpha_i$ and $\beta_i$
respectively using $\R^M$ vectors $\vec{w}$, $\vec{\alpha}$, and $\vec{\beta}$. 

The cross-entropy loss function per instance including the L2 regularization terms for
the weight vector $\vec{w}$ and scale vector $\vec{\beta}$ can be written as follows:
\begin{eqnarray}
\label{eq:loss}
L(\vec{w},b,\vec{\alpha},\vec{\beta}) = -t_n \log y_n - (1-t_n) \log (1-y_n) 
\end{eqnarray}
Here, we used $y_n = P(t=1 | \vec{x}^n, b, \vec{\alpha}, \vec{\beta})$ to minimize the cluttering
of symbols in Equation \ref{eq:loss}. 
To avoid overfitting to training instances and to minimize the distortion of the training instances,
we impose L2 regularization on $\vec{w}$, $\vec{\alpha}$, and $\vec{\beta}$. 
Therefore, the final objective function that must be minimized with respect to
$\vec{w}$, $\vec{\alpha}$, $\vec{\beta}$, and $b$ is give by,
\begin{eqnarray}
\label{eq:total}
E(\vec{w},b,\vec{\alpha},\vec{\beta}) = L(\vec{w},b,\vec{\alpha},\vec{\beta}) + \lambda \norm{w}_2^2 + \mu \norm{\alpha}_2^2 + \nu \norm{\beta}_2^2
\end{eqnarray}

Here, $\lambda$, $\mu$ and $\nu$ respectively are the L2 regularization coefficients corresponding to
the weight vector $\vec{w}$ and the scale vectors $\vec{\alpha}$, $\vec{\beta}$. 
Because we consider the minimization of Equation \ref{eq:total} per instance basis,
in our experiments, we divide the regularization parameters $\lambda$, $\mu$, and $\nu$ by the
total number of training instances $N$ in the dataset such that we can compare the values
those parameters across datasets of different sizes.

By setting the partial derivatives $\frac{\partial E}{\partial w_j}$, $\frac{\partial E}{\partial b}$, 
$\frac{\partial E}{\partial \alpha_j}$, and $\frac{\partial E}{\partial \beta_j}$
to zero and applying Stochastic Gradient Descent (SGD) update rule the following updates can be derived,
\begin{eqnarray}
\label{eq:w_update}
w_j^{k+1} = w_j^{k}(1 - 2 \lambda \eta_k) + \eta_k (t_n - y_n) \sigma_j(x^n_j) ,
\end{eqnarray}
\begin{eqnarray}
\label{eq:b_update}
b^{k+1} = b^{k}+ \eta_k(t_n - y_n) ,
\end{eqnarray}
\begin{eqnarray}
\label{eq:alpha_update}
\alpha_j^{k+1} &= \alpha_j^k (1 - 2 \mu \eta_k) + \eta_k x^n_j w_j \sigma_j(x^n_j) (1 - \sigma_j(x^n_j)) (t_n - y_n) ,
\end{eqnarray}
\begin{eqnarray}
\label{eq:beta_update}
\beta_j^{k+1} &= \beta_j^k (1 - 2 \nu \eta_k) - \eta_k (t_n - y_n) w_j \sigma_j(x^n_j) (1 - \sigma_j(x^n_j)) .
\label{eq:beta_update} 
\end{eqnarray}

In Equations \ref{eq:w_update}-\ref{eq:beta_update}, $k$ denotes the $k$-th update and 
$\eta_k$ is the learning rate for the $k$-th update.
We experimented with both linear and exponential decaying and found linear decaying to perform
better for the proposed method. The linear decaying function for $\eta_k$ is defined as follows,
\begin{equation}
\eta_k = \frac{\eta_0}{1 + \frac{k}{T \times N}} .
\end{equation}
Here, $T$ is the total number of iterations for which the training dataset containing $N$ instances
will be traversed. Because we are considering OPOL, we set $T=1$. The initial learning rate
$\eta_0$ is set to $0.1$ throughout the experiments described in the paper.
This value of $0.1$ was found to be producing the best results in our preliminary experiments
using development data, which we selected randomly from the benchmark datasets described
later in Section \ref{sec:datasets}.

Several observations are in order. First, note that the scaling factors $\alpha_j$
and $\beta_j$ distort the original value of the feature $x_i$.
If this distortion is too much, then we might loose the information conveyed by the
feature $x_i$. To minimize the distortion of $\vec{x}$ because of
scaling, we have imposed regularization on both $\vec{\alpha}$ and $\vec{\beta}$.
This treatment is similar to the slack variables often used in non-separable
classification tasks and imposing a penalty on the total slackness. 
Of course, the regularization on $\vec{\alpha}$ and $\vec{\beta}$ can be removed
simply by setting the corresponding regularization coefficients $\mu$ and $\nu$
to zero. Therefore, the introduction of regularization on $\vec{\alpha}$
and $\vec{\beta}$ does not harm the generality of the proposed method.
The total number of parameters to train in this model is $M + M + M + 1 = 3M + 1$,
corresponding to $\vec{w}$, $\vec{\alpha}$, $\vec{\beta}$, and $b$.
Note that we must not regularize the bias term $b$ and let it to adjust arbitrarily.
This can be seen as a dynamic scaling for the score (i.e. inner-product between $\vec{w}$ and $\vec{x}$),
although this type of scaling is \textit{not} feature specific.  
The sigmoid-based feature scaling function given by Equation \ref{eq:trans} is by no means the only
function that satisfies the requirement for a scaling function (i.e. maps all feature values
to the same range such as $[0,1]$). However, the sigmoid function has been widely used in various fields
of machine learning such as neural networks \cite{Zhang:2000}, and has desirable  properties such as differentiability
and continuity. 

Next, we introduce several important variants of Equation \ref{eq:trans} and present the
update equations for each of those variants. In Section \ref{sec:exps}, we empirically study the
effect of the different variants discussed in the paper.
For the ease of reference, we name the original formulation given by Equation \ref{eq:trans}
as \textbf{FS} (Supervised Feature Scaling) method.
The objective function given by Equation \ref{eq:total} is convex with respect to 
$\vec{w}$. This can be easily verified by computing the second derivative of the
objective function with respect to $w_i$, which becomes
\begin{equation}
\label{eq:w_d2}
\frac{\partial^2 E}{\partial w_i^2} = {\sigma(x_i)}^2 y_n (1 - y_n) + 2\lambda .
\end{equation}
Because $0 < \sigma(x_i) < 1$, $0 < y_n < 1$, and $0 < \lambda$ hold,
the second derivative $\frac{\partial^2 E}{\partial w_i^2} > 0$,
which proves that the objective function is convex with respect to $w_i$.
Likewise, the objective function can be shown to be convex with respect to the bias term $b$.
It is interesting to note that the convexity holds irrespective of the form of the
scaling function $\sigma$ for both $\vec{w}$ and $b$ as long as $\sigma(x_i) \neq 0$ is satisfied.
If $\sigma(x_i) = 0$ for some value of $x_i$, then the convexity of $E$ also
depends upon $\lambda$ not being equal to zero. 
Although, in the case of sigmoid feature scaling functions $\sigma(x_i) \to 0$ when $x_i \to -\infty$
this is irrelevant because feature values are finite in practice. 
Unfortunately, the objective function is \textit{non-convex} with respect to
$\vec{\alpha}$ and $\vec{\beta}$. Although SGD updates are empirically shown to work well even  
when the objective function is non-convex, there is no guarantee that the update
Equations \ref{eq:w_update} - \ref{eq:beta_update} will find the global minimum of the
objective function.

\subsection{\textbf{FS-1}}
\label{sec:FS1}

In this variant we fix the scaling factor $\vec{\alpha} = \vec{1}$, thereby reducing the
number of parameters to be tuned. However, this model cannot adjust for the different
value ranges of features and can only learn the shiftings required.
We name this variant as \textbf{FS-1} and is given by,
\begin{equation}
\label{eq:FS1}
\sigma_i(x_i^n) = \frac{1}{1+\exp(-x_i^n + \beta_i)}. 
\end{equation}
The update equations for $w_j$, $b$, and $\beta_j$ are as follows,
\begin{eqnarray}
\label{eq:FS1:w_update}
w_j^{k+1} = w_j^{k}(1 - 2 \lambda \eta_k) + \eta_k (t_n - y_n) \sigma_j(x^n_j) ,
\end{eqnarray}
\begin{eqnarray}
\label{eq:FS1:b_update}
b^{k+1} = b^{k}+ \eta_k(t_n - y_n) ,
\end{eqnarray}
\begin{eqnarray}
\label{eq:FS1:beta_update}
\beta_j^{k+1} &= \beta_j^k (1 - 2 \nu \eta_k) - \eta_k (t_n - y_n) w_j \sigma_j(x^n_j) (1 - \sigma_j(x^n_j)) . 
\end{eqnarray}
Note that although the update Equations \ref{eq:FS1:w_update}, \ref{eq:FS1:b_update}, and \ref{eq:FS1:beta_update}
appear to be similar in their form to Equations \ref{eq:w_update}, \ref{eq:b_update}, and \ref{eq:beta_update},
the transformation functions in the two sets of equations are different.
As discussed earlier under \textbf{FS}, \textbf{FS-1} is also convex with respect to $\vec{w}$ and $b$, but non-convex
with respect to $\vec{\beta}$.

\subsection{\textbf{FS-2}}
\label{sec:FS2}

We design a convex form of the objective function with respect to all parameters by replacing the
sigmoid feature scaling function with a linear combination as follows,
\begin{equation}
\label{eq:sigma_conex}
\sigma_i(x_i) = \alpha_i x_i + \beta_i .
\end{equation}
The class conditional probability is computed using the logistic sigmoid model as,
\begin{equation}
\label{eq:convex_sigmoid}
P(t_n = 1 | \vec{w}, b, \vec{\alpha}, \vec{\beta}) = \frac{1}{1 + \exp(-\sum_{j=1}^{M} w_j (\alpha_j x^n_j + \beta_j) -b)} .
\end{equation}

Then the update equations for $\vec{w}$, $b$, $\vec{\alpha}$, and $\vec{\beta}$ are given as follows, 
\begin{eqnarray}
\label{eq:FS2:w_update}
w^{k+1}_j = w^{k}_j (1 - 2 \lambda \eta_k) - \eta_k (y_n - t_n) (\alpha_j x^n_j + \beta_j) ,
\end{eqnarray}
\begin{eqnarray}
\label{eq:FS2:b_update}
b^{k+1} = b^k - \eta_k (y_n - t_n),
\end{eqnarray}
\begin{eqnarray}
\label{eq:FS2:alpha_update}
\alpha^{k+1}_j = \alpha^k_j (1 - 2 \mu \eta_k) - \eta_k (y_n - t_n) w_j x^n_j,
\end{eqnarray}
\begin{eqnarray}
\label{eq:FS2:beta_update}
\beta^{k+1}_j = \beta^k_j (1 - 2 \nu \eta_k) - \eta_k (y_n - t_n) w_j .
\end{eqnarray}
Here, we used $y_n = P(t_n = 1 | \vec{w}, b, \vec{\alpha}, \vec{\beta})$ to simplify the equations.

Moreover, the second-order partial derivatives of the objective function $E$, with respect to
$\vec{w}$, $b$, $\vec{\alpha}$, and $\vec{\beta}$ can be computed as follows,
\begin{eqnarray}
\nonumber
\frac{\partial^2 E}{\partial w_j^2} &=& y_n (1 - y_n) {(\alpha_j x^n_j + \beta_j)}^{2} + 2 \lambda, \\ \nonumber
\frac{\partial^2 E}{\partial \alpha_j^2} &=& y_n (1 - y_n) w_j^2 x^{n2}_j + 2 \mu, \\ \nonumber
\frac{\partial^2 E}{\partial \beta_j^2} &=& y_n (1 - y_n) w_j^2 x^{n2}_j + 2 \mu, \\ \nonumber
\frac{\partial^2 E}{\partial w_j^2} &=& y_n (1 - y_n) .
\end{eqnarray}
From, $0 < y_n < 1$, $\lambda > 0$, $\mu > 0$, and $\nu > 0$ it follows that all of the above-mentioned
second-order derivatives are positive, which proofs the convexity of the objective function.
We name this convex formulation of the feature scaling method as the \textbf{FS-2} method.

\subsection{\textbf{FS-3}}
\label{sec:FS3}

Although \textbf{FS-2} is convex, there is an issue regarding the determinability among
$\vec{w}$, $\vec{\alpha}$, and $\vec{\beta}$ because the product between $\vec{w}$ and $\vec{\alpha}$,
and the product between $\vec{w}$ and $\vec{\beta}$ appear inside the exponential term
in Equation \ref{eq:convex_sigmoid}. This implies that the probability $P(t_n = 1 | \vec{w}, b, \vec{\alpha}, \vec{\beta})$ will be invariant under a constant scaling of $\vec{w}$, $\vec{\alpha}$, and $\vec{\beta}$.
We can absorb the $w_j$ terms from the objective function into the corresponding $\alpha_j$
and $\beta_j$ terms thereby effectively both reducing the number of parameters to be trained
as well as eliminating the issue regarding the determinability. 
We name this variant of the feature scaling method as the \textbf{FS-3} method.

The class conditional probability for \textbf{FS-3} is give by,
\begin{equation}
P(t_n = 1 | b, \vec{\alpha}, \vec{\beta}) =  \frac{1}{1 + \exp(-\sum_{j=1}^{M} (\alpha_j x^n_j + \beta_j) -b)} .
\end{equation}
This can be seen as a special case of \textbf{FS-2} where we set $\vec{w} = \vec{1}$ and $\lambda = 0$.

The update equations for \textbf{FS-3} can be derived as follows,
\begin{eqnarray}
\label{eq:FS3:b_update}
b^{k+1} = b^k - \eta_k (y_n - t_n),
\end{eqnarray}
\begin{eqnarray}
\label{eq:FS3:alpha_update}
\alpha^{k+1}_j = \alpha^k_j (1 - 2 \mu \eta_k) - \eta_k (y_n - t_n) x^n_j,
\end{eqnarray}
\begin{eqnarray}
\label{eq:FS3:beta_update}
\beta^{k+1}_j = \beta^k_j (1 - 2 \nu \eta_k) - \eta_k (y_n - t_n) .
\end{eqnarray}
Here, we used $y_n = P(t_n = 1 | b, \vec{\alpha}, \vec{\beta})$ to simplify the equations.
Because \textbf{FS-2} is convex and \textbf{FS-3} is a special case of \textbf{FS-2},
it follows that \textbf{FS-3} is also convex.

\section{Datasets}
\label{sec:datasets}

To evaluate the performance of the numerous feature scaling methods introduced in Section \ref{sec:supscale},
we train and test those methods under the one-pass online learning setting. We use three datasets
in our experiments: \textbf{heart} dataset, \textbf{liver} dataset, and the \textbf{diabetes} dataset.
All three datasets are popularly used as benchmark datasets to evaluate binary classification 
algorithms. Moreover, all three datasets contain real-valued and unscaled features values,
which are appropriate for the current evaluation purpose. All three datasets can be downloaded from 
the UCI Machine Learning Repository\footnote{\url{http://archive.ics.uci.edu/ml/}}.
Details of the three datasets are summarized in Table \ref{tbl:datasets}.

\begin{table}[h]
\caption{Statistics regarding the three datasets used in the experiments.}
\begin{center}
\begin{tabular}{|l||c|c|c|}\hline
Dataset & Attributes & Train instances & Test instances \\ \hline \hline
heart & $13$ & $216$ & $54$ \\ \hline
liver & $6$ & $276$ & $69$ \\ \hline
diabetes & $8$ & $611$ & $157$ \\ \hline 
\end{tabular}
\end{center}
\label{tbl:datasets}
\end{table}

\section{Experiments and Results}
\label{sec:exps}

To compare the performance of the different dynamic feature scaling methods we proposed in
the paper, we use those methods to scale features in the following online learning algorithms.
\begin{description}
\itemsep 10pt
\item[SGD (Stochastic Gradient Descent):] 
This method implements logistic regression using stochastic gradient descent.
It does not use any feature scaling and uses the original feature values as they are
for training a binary classifier. This method demonstrates the lower baseline performance
for this task.

\item[SDG+avg (Stochastic Gradient Descent with Model Averaging):]
This method is the same as \textbf{SGD} described above, except that it uses the average 
weight vector during training and testing. Specifically, it computes the average of the
weight vector $\vec{w}$ over the updates and uses this average vector for prediction.
By considering the average weight vector instead of the final weight vector we can
avoid any bias toward the last few training instances encountered by the online learner.
Moreover, it has been shown both theoretically and empirically that consideration of the average weight
vector results in faster convergence in online learning \cite{Collins:EMNLP:2002}.

\item[GN (Unsupervised Dynamic Scaling):]
This is the unsupervised dynamitc feature scaling method described in Section \ref{sec:unsupscale}.
It trains a binary logistic regression model by scaling the features using the unsupervised
approach.

\item[GN+avg (Unsupervised Dynamic Scaling with Model Averaging):]
This is the unsupervised feature scaling method described in Section \ref{sec:unsupscale}
using the average weight vector for predicting instead of the final weight vector.
It trains a binary logistic regression model by scaling the features using the unsupervised
approach. 

\item[FS (Supervised Dynamic Feature Scaling):]
This is the supervised dynamic feature scaling method described in Section \ref{sec:supscale}.

\item[FS+avg (Supervised Dynamic Feature Scaling with Model Averaging):]
This is the \textbf{FS} method, where we use the average values for all parameters:
$\vec{w}$, $b$, $\vec{\alpha}$, and $\vec{\beta}$.

\item[FS-1 (Supervised Dynamic Feature Scaling variant FS-1):]
This is the method described in Section \ref{sec:FS1}.

\item[FS-1+avg (Supervised Dynamic Feature Scaling variant FS-1 with Model Averaging):]
This is the method described in Section \ref{sec:FS1} with averaged parameter vectors.

\item[FS-2 (Supervised Dynamic Feature Scaling variant FS-2):]
This is the method described in Section \ref{sec:FS2}.

\item[FS-2+avg (Supervised Dynamic Feature Scaling variant FS-1 with Model Averaging):]
This is the method described in Section \ref{sec:FS2} with averaged parameter vectors.

\item[FS-3 (Supervised Dynamic Feature Scaling variant FS-3):]
This is the method described in Section \ref{sec:FS3}.

\item[FS-3+avg (Supervised Dynamic Feature Scaling variant FS-1 with Model Averaging):]
This is the method described in Section \ref{sec:FS3} with averaged parameter vectors.

\item[PA (Passive-Aggressive):]
This is the Passive-Aggressive binary linear classification algorithm proposed by \cite{Crammer:2006}.

\item[PA+avg (Passive-Aggressive with Model Averaging):]
This is the Passive-Aggressive binary linear classification algorithm proposed by \cite{Crammer:2006}
using the averaged weight vector to predict during both training and testing stages.

\item[PA-1 (Passive-Average variant 1):]
This is the Passive-Aggressive PA-I version of the binary 
linear classification algorithm proposed by \cite{Crammer:2006}.

\item[PA-1+avg (Passive-Aggressive variant 1 with Model Averaging):]
This is the Passive-Aggressive PA-1 version of the binary 
linear classification algorithm proposed by \cite{Crammer:2006} using 
the averaged weight vector to predict during both training and testing stages.

\item[PA-2 (Passive-Aggressive variant 2):]
This is the Passive-Aggressive PA-2 version of the binary 
linear classification algorithm proposed by \cite{Crammer:2006}.

\item[PA-2+avg (Passive-Aggressive variant 2 with Model Averaging):]
This is the Passive-Aggressive PA-2 version of the binary 
linear classification algorithm proposed by \cite{Crammer:2006} using 
the averaged weight vector to predict during both training and testing stages.

\end{description}

\begin{table}[t]
\caption{Results on the heart dataset.}
\begin{center}
\begin{tabular}{|l||c|c|c|}\hline
Algorithm	& Train Accuracy &	Test Accuracy & Best Parameters 	\\ \hline \hline
SGD	& $0.537037$ &	$0.574074$ &	 $\lambda=0.01$ \\
SGD+avg &	$0.481481$ &	$0.435185$ &	 $\lambda=0$ \\
\textbf{GN}	& $\mathbf{0.87037}$ &	$\mathbf{0.824074}$ &	 $\lambda=0.01$ \\
GN+avg	& $0.777778$ &	$0.768519$ &	 $\lambda=0.1$ \\
FS	& $0.592593$ &	$0.49537$ &	 $\lambda=0.1, \mu=1.0, \nu=0$ \\
FS+avg &	$0.481481$ &	$0.435185$ &	 $\lambda=0, \mu=0 \nu=0$ \\
FS-1 &	$0.703704$	& $0.564815$ &	$\mu=100.0, \nu=0.1$ \\
FS-1+avg &	$0.759259$ &	$0.564815$ &	 $\mu=0.1, \nu=10.0$ \\
FS-2 & 	$0.740741$ &	$0.569444$ &	 $\lambda=10.0, \mu=0, \nu=10.0$ \\
FS-2+avg &	$0.574074$ &	$0.467593$ &	$\lambda=0, \mu=1.0, \nu=0$ \\
FS-3 &	$0.592593$ &	$0.476852$	& $\mu=0.1, \nu=0$ \\
FS-3+avg & 	$0.574074$ & 	$0.421296$ &	 $\mu=0.1, \nu=1.0$ \\
PA	& $0.648148$ &	$0.675926$ &	 $c=0.01$ \\
PA+avg	& $0.611111$ &	$0.662037$ &	 $c=0.01$ \\
PA1	& $0.648148$ &	$0.675926$ &	 $c=0.01$ \\
PA1+avg &	$0.611111$ &	$0.662037$ &	 $c=0.01$ \\
PA2	& $0.648148$ &	$0.675926$ &	 $c=0.01$ \\
PA2+avg	& $0.611111$ &	$0.662037$ &	 $c=0.01$ \\ \hline
\end{tabular}
\end{center}
\label{tbl:heart}
\end{table}

\begin{table}[t]
\caption{Results on the liver dataset.}
\begin{center}
\begin{tabular}{|l||c|c|c|}\hline
Algorithm	& Train Accuracy &	Test Accuracy & Best Parameters 	\\ \hline \hline
SGD	& $0.608696$ &	$0.561594$ &	 $\lambda=0.1$ \\
SGD+avg &	$0.550725$ &	$0.586957$ &	 $\lambda=0$ \\
GN	& $0.695652$ &	$0.637681$ &	 $\lambda=100.0$ \\
\textbf{GN+avg}	& $\mathbf{0.777778}$ &	$\mathbf{0.768519}$ &	 $\lambda=0.1$ \\
FS	& $0.637681$ &	$0.586957$ &	 $\lambda=10.0, \mu=0.1, \nu=0.1$ \\
FS+avg &	$0.550725$ &	$0.586957$ &	 $\lambda=0, \mu=0 \nu=0$ \\
FS-1 &	$0.623188$	& $0.413043$ &	$\mu=1.0, \nu=0$ \\
FS-1+avg &	$0.623188$ &	$0.413043$ &	 $\mu=0.1, \nu=0.1$ \\
FS-2 & 	$0.681159$ &	$0.59058$ &	 $\lambda=0, \mu=0.01, \nu=0$ \\
FS-2+avg &	$0.550725$ &	$0.586957$ &	$\lambda=0, \mu=0, \nu=0$ \\
FS-3 &	$0.623188$ &	$0.550725$	& $\mu=0, \nu=0$ \\
FS-3+avg & 	$0.550725$ & 	$0.586957$ &	 $\mu=0, \nu=0$ \\
PA	& $0.434783$ &	$0.427536$ &	 $c=0.01$ \\
PA+avg	& $0.565217$ &	$0.594203$ &	 $c=0.01$ \\
PA1	& $0.434783$ &	$0.427536$ &	 $c=0.01$ \\
PA1+avg &	$0.565217$ &	$0.594203$ &	 $c=0.01$ \\
PA2	& $0.434783$ &	$0.427536$ &	 $c=0.01$ \\
PA2+avg	& $0.565217$ &	$0.594203$ &	 $c=0.01$ \\ \hline
\end{tabular}
\end{center}
\label{tbl:liver}
\end{table}

\begin{table}[t]
\caption{Results on the diabetes dataset.}
\begin{center}
\begin{tabular}{|l||c|c|c|}\hline
Algorithm	& Train Accuracy &	Test Accuracy & Best Parameters 	\\ \hline \hline
SGD	&	$0.643312$	& $0.653028$ &	 $\lambda=1.0$ \\
SGD+avg &	$0.643312$ &	$0.653028$ &	 $\lambda=0$ \\
GN	& $0.656051$ &	$0.656301$ &	 $\lambda=0.01$ \\
\textbf{GN+avg}	& $\mathbf{0.656051}$ &	$\mathbf{0.671031}$ &	 $\lambda=100.0$ \\
FS	& $0.643312$ &	$0.653028$ &	 $\lambda=0, \mu=0, \nu=0$ \\
FS+avg &	$0.643312$ &	$0.653028$ &	 $\lambda=0, \mu=0 \nu=0$ \\
FS-1 &	$0.643312$	& $0.653028$ &	$\mu=10, \nu=0$ \\
FS-1+avg &	$0.643312$ &	$0.653028$ &	 $\mu=0, \nu=0$ \\
FS-2 & 	$0.643312$ &	$0.653028$ &	 $\lambda=0, \mu=0, \nu=1.0$ \\
FS-2+avg &	$0.643312$ &	$0.653028$ &	$\lambda=0, \mu=0, \nu=0$ \\
FS-3 &	$0.643312$ &	$0.653028$	& $\mu=0.01, \nu=100.0$ \\
FS-3+avg & 	$0.643312$ & 	$0.653028$ &	 $\mu=0, \nu=0.01$ \\
PA	& $0.611465$ &	$0.656301$ &	 $c=0.01$ \\
PA+avg	& $0.636943$ &	$0.657938$ &	 $c=0.01$ \\
PA1	& $0.648148$ &	$0.675926$ &	 $c=0.01$ \\
PA1+avg &	$0.636943$ &	$0.657938$ &	 $c=0.01$ \\
PA2	& $0.611465$ &	$0.656301$ &	 $c=0.01$ \\
PA2+avg	& $0.636943$ &	$0.657938$ &	 $c=0.01$ \\ \hline
\end{tabular}
\end{center}
\label{tbl:diabetes}
\end{table}

We measure train and test classification accuracy for each of the above-mentioned $18$ algorithms.
Classification accuracy is defined as follows:
\begin{equation}
\label{eq:error}
\mbox{Classification Accuracy} = \frac{\mbox{total no. of correctly classified instances}}
{\mbox{total no. of instances in the dataset}} .
\end{equation}
Note that all three benchmark datasets described in Section \ref{sec:datasets} are balanced 
(i.e. contains equal numbers of positive and negative train/test instances).
Therefore, a method that randomly classifies test instances would obtain an accuracy of $0.5$.
 The experimental results for heart, liver, and diabetes datasets are shown respectively in Tables \ref{tbl:heart},
\ref{tbl:liver}, and \ref{tbl:diabetes}.

We vary the values for the numerous parameters in a pre-defined set of values for each parameter
and experiment with all possible combinations of those values. For the regularization
coefficients $\lambda$, $\mu$, and $\nu$ we experiment with the values in the set
$\{0, 0.01, 0.1, 1, 10, 100\}$. For the $c$ parameter in passive-aggressive algorithms we
chose from the set $\{0.01, 0.1, 1, 10, 100\}$. 
In each dataset, we randomly set aside $1/5$-th of all training data for validation purposes. 
We search for the parameter values for each algorithm that produces the highest accuracy on
the validation dataset. Next, we fix those parameter values and evaluate on the test portion 
of the corresponding dataset. The best parameter values found through the search procedure are
shown in the fourth column in Tables \ref{tbl:heart}-\ref{tbl:diabetes}.
Online learning algorithms have been shown to be sensitive to the order in which
training examples are presented to them. Following the suggestions in prior work,
we randomize the sequence of training data instances during training \cite{Berrtsekas:2010}.
All results shown in the paper are the average of $10$ random initializations.

As can be seen from Tables \ref{tbl:heart}, \ref{tbl:liver}, and \ref{tbl:diabetes}
the unsupervised dynamic feature scaling methods (\textbf{GN} and \textbf{GN+avg}) consistently
outperform joint supervised feature scaling methods and PA algorithms.
Model averaged version of the unsupervised dynamic feature scaling method (\textbf{GN+avg})
shows better performance than its counterpart that does not perform model averaging (\textbf{GN})
in two out of the tree datasets. Compared to the unsupervised dynamic feature scaling methods 
(\textbf{GN} and \textbf{GN+avg}), the supervised dynamic feature scaling methods 
(\textbf{FS}, \textbf{FS-1}, \textbf{FS-2}, and \textbf{FS-3}) report lower test accuracies.
Compared to the unsupervised dynamic feature scaling methods, the number of parameters that must be estimated
from labeled data is larger in the supervised dynamic feature scaling methods. 
Although the unsupervised dynamic feature scaling method requires us to estimate the
mean and standard deviation from train data, those parameters can be estimated without using the
label information in the training instances.
Therefore, the unsupervised dynamic feature scaling is less likely to overfit to the train data,
which results in better performance. 

Recall that \textbf{SGD} and \textbf{SGD+avg} do not perform any dynamic feature scaling and
demonstrate the level of accuracy that we would obtain if we had not perform feature scaling.
In all datasets, the \textbf{GN} and \textbf{GN+avg} methods significantly outperform 
(according to a two-tailed paired t-test under $0.05$ confidence level) the SGD counterparts
showing the effectiveness of feature scaling when training binary classifiers.

Among the variants of the proposed \textbf{FS} methods, the \textbf{FS-2} method reports
the best performance. We believe that this can be attributable to the convexity of the objective function.
Because we are allowed only a single pass over the training dataset in OPOL setting,
convergence becomes a critical issue compared to the classical online learning setting
where the learning algorithm traverses multiple times over the dataset. 
Convex functions can be relatively easily optimized using gradient methods compared to
non-convex functions. \textbf{FS-3} method which constrains the parameters in the \textbf{FS-2}
method shows poor performance in our experiments. Specifically, \textbf{FS-3} absorbs the
weight parameters into the scaling parameters in the \textbf{FS-2} method.
However, the experimental results show that we should keep the two sets of parameters separately.
In our future work, we plan to study other possible ways to reduce the number of parameters
in the supervised dynamic feature scaling methods in order to reduce the effect of overfitting.
Among the three datasets, the performance differences of the methods compared are least
significant on the diabetes dataset. In fact, $10$ of the $18$ methods report the same test
accuracy on this dataset and learns the same classification model. However, the model averaged 
version of the unsupervised dynamic feature scaling method (\textbf{GN+avg}) outperforms all
the methods compared even in the diabetes dataset that shows its ability to perform well even
in situations where other methods cannot.

\begin{figure}[t]
\begin{center}
\includegraphics[width=90mm]{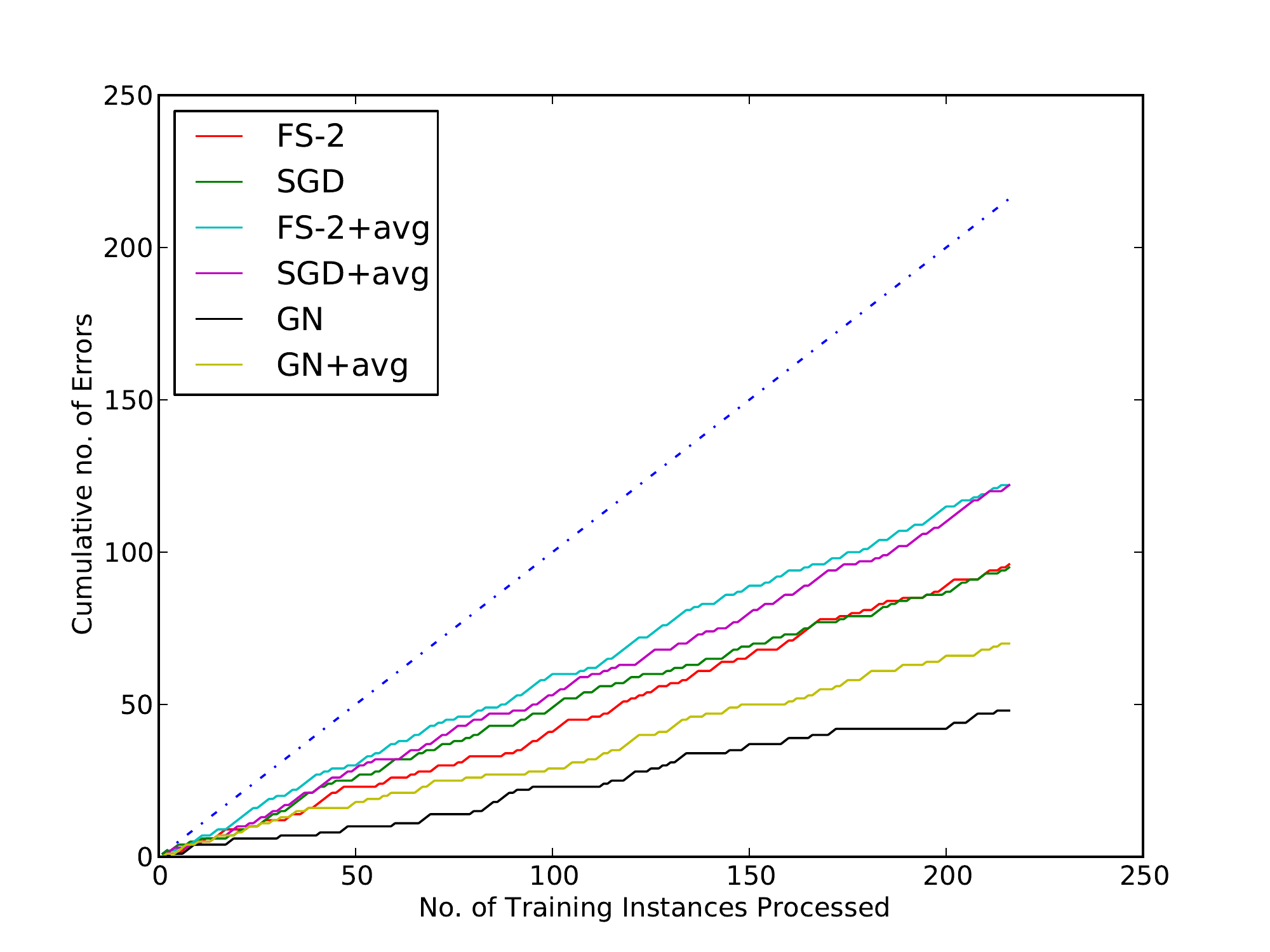}
\caption{Cumulative training errors on the heart dataset.}
\label{fig:cum_heart}
\end{center}
\end{figure}

\begin{figure}[t]
\begin{center}
\includegraphics[width=90mm]{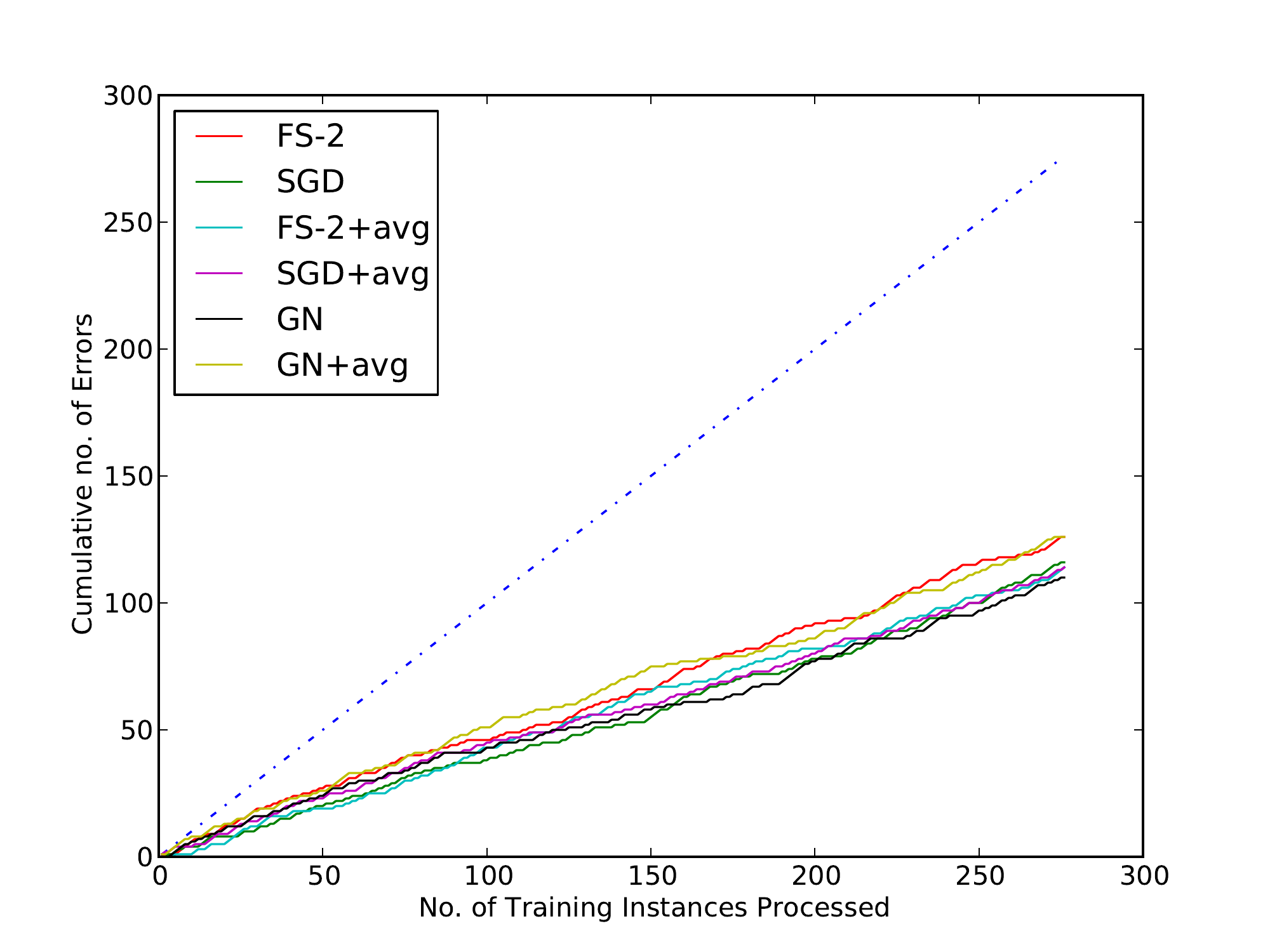}
\caption{Cumulative training errors on the liver dataset.}
\label{fig:cum_liver}
\end{center}
\end{figure}

To study the behavior of the different learning algorithms during train time, we
compute the cumulative number of errors. Cumulative number of errors represents 
the total misclassification errors encountered up to the current train instance. 
In an one-pass online learning setting, we must continuously both train as well as apply the trained
classifier to classify new instances on the fly. Therefore, a method that obtains a lower
number of cumulative errors is desirable. To compare the different methods described in the paper,
we plot the cumulative number of errors against the total number of training instances encountered as shown in 
Figures \ref{fig:cum_heart} and \ref{fig:cum_liver},
respectively for heart and liver datasets. 
During training, we use the weight vector (or the averaged weight vector for the \textbf{+avg} methods)
to classify the current training instances and if it is misclassified by the
current model, then it is counted as an error.
The 45 degree line in each plot corresponds to the situation where all instances
encountered during training are misclassified. All algorithms must lie below this line.
To avoid cluttering, we only show the cumulative number of error curves for
the following six methods: \textbf{FS-2}, \textbf{FS-2+avg}, \textbf{SGD},
\textbf{SGD+avg}, \textbf{GN}, and \textbf{GN+avg}.
Overall, we see that the unsupervised dynamic feature scaling methods \textbf{GN} and
\textbf{GN+avg} stand out among the others and report lower numbers of cumulative errors.

\section{Conclusion}
\label{sec:conclusion}

We studied the problem of feature scaling in one-pass online learning (OPOL) of binary linear classifiers.
In OPOL, a learner is allowed to traverse a dataset only once. 
We presented both supervised as well as unsupervised approaches to dynamically scale feature
under the OPOL setting. We evaluated $18$ different learning methods using three popular datasets. 
Our experimental results show that 
the unsupervised approach significantly outperforms the supervised approaches and improves the 
classification accuracy in a state-of-the-art online learning algorithm proposed by \cite{Crammer:2006}.
Among the several variants of the supervised feature scaling approach we evaluated,
the convex formulation performed best.
In future, we plan to explore other forms of feature scaling functions and their effectiveness 
in numerous online learning algorithms proposed for classification.

\bibliographystyle{plain}
\bibliography{featscale}
\end{document}